\documentclass[letterpaper, 11pt]{criarticle}
\usepackage{amsmath} 
\usepackage{amsthm} 
\usepackage{amsfonts} 
\usepackage{amssymb}
\usepackage{multirow}
\usepackage{dcolumn}
\usepackage{array}
\usepackage{graphicx}
\usepackage{algorithm}
\usepackage[noend]{algpseudocode}

\usepackage{graphicx} 
\usepackage{comment}
\usepackage[inline]{enumitem}
\usepackage{flushend}
\usepackage{amsmath}
\usepackage{amssymb}
\usepackage{colortbl}
\usepackage{cellspace}
\usepackage{xcolor}
\usepackage{booktabs}
\usepackage[numbers]{natbib}

\usepackage{titlesec}
\usepackage{enumitem}
\usepackage{url}
\usepackage{authblk}
\usepackage{graphicx}
\usepackage{caption}
\usepackage{booktabs}

\usepackage{tikz}
\usepackage{pgfplots}
\usepgfplotslibrary{units}
\usepackage[T1]{fontenc}
\usepackage{txfonts}
\usepackage{fixltx2e}
\usepackage{tabu}

\newcolumntype{C}[1]{>{\centering\let\newline\\\arraybackslash\hspace{0pt}}m{#1}}

\usepackage[margin=0.75in]{geometry}
\usepackage{color,hyperref}
\definecolor{darkblue}{rgb}{0.0,0.0,0.3}
\hypersetup{colorlinks,breaklinks,
linkcolor=darkblue,urlcolor=darkblue,
anchorcolor=darkblue,citecolor=darkblue}
\usepackage{url}
\usepackage{graphicx}
\usepackage[mathscr]{euscript}
\usepackage{mathtools}
\DeclarePairedDelimiterX{\norm}[1]{\lVert}{\rVert}{#1}
\usepackage{comment}

\date{}
\newlist{myenumi}{description}{10}
\setlist[myenumi]{labelindent=\parindent, leftmargin=*, label=(\arabic*), align=left}
\setlist[myenumi]{leftmargin=15pt}

\usepackage{setspace}
\begin{document}

\begin{titlepage}

  \newcommand{\HRule}{\rule{\linewidth}{1.2mm}} 

  \center 




  \HRule \\[0.4cm]
  { \Large \bfseries A Marker-based Neural Network System for Extracting\\Social 
Determinants of Health}\\[0.3cm] 
  \HRule \\[0.5cm]
 

  {\em\Large Authors}\\
  \vspace{.3 cm}
  \textbf{Xingmeng Zhao}\\
  Information Systems and Cyber Security\\
  University of Texas at San Antonio, USA\\
  
  \vspace{.2 cm}
  \textbf{Anthony  Rios}, Ph.D\\
  Information Systems and Cyber Security\\
  University of Texas at San Antonio, USA\\
  

  \vspace{0.3 cm}
  \begin{flushleft}
  {\em\large{\textbf{Corresponding Author}}: Anthony Rios,  \\
  Email: anthony.rios@utsa.edu}\\
  \vspace{0.2 cm}
 \vspace{0.2 cm}
  {\em\large \textbf{Word Count}: ~3936}
  \end{flushleft}

\vfill 

\end{titlepage}


\newpage
\title{\Large A Marker-based Neural Network System for Extracting Social 
Determinants of Health}
\author{Xingmeng Zhao}
\author{Anthony Rios, Ph.D.}
\affil{Department of Information Systems and Cyber Security, University of Texas at San Antonio, USA}

\newcommand\Mycite[1]{%
  \citeauthor{#1}~(\citeyear{#1})}

\maketitle
\thispagestyle{empty}

\begin{abstract}

\noindent \textbf{Objective.} The impact of social determinants of health (SDoH) on patients' healthcare quality and the disparity is well-known. Many SDoH items are not coded in structured forms in electronic health records. These items are often captured in free-text clinical notes, but there are limited methods for automatically extracting them. We explore a multi-stage pipeline involving named entity recognition (NER), relation classification (RC), and text classification methods to extract SDoH information from clinical notes automatically.

\vspace{2mm}
\noindent \textbf{Materials and Methods.} The study uses the N2C2 Shared Task data, which was collected from two sources of clinical notes: MIMIC-III and University of Washington Harborview Medical Centers. It contains 4480 social history sections with full annotation for twelve SDoHs. In order to handle the issue of overlapping entities, we developed a novel marker-based NER model. We used it in a multi-stage pipeline to extract SDoH information from clinical notes.

\vspace{2mm}
\noindent \textbf{Results.} Our marker-based system outperformed the state-of-the-art span-based models at handling overlapping entities based on the overall Micro-F1 score performance. It also achieved state-of-the-art performance compared to the shared task methods.

\vspace{2mm}
\noindent \textbf{Conclusion.} The major finding of this study is that the multi-stage pipeline effectively extracts SDoH information from clinical notes. This approach can potentially improve the understanding and tracking of SDoHs in clinical settings. However, error propagation may be an issue, and further research is needed to improve the extraction of entities with complex semantic meanings and low-resource entities using external knowledge.

\end{abstract}

\section{BACKGROUND AND SIGNIFICANCE}
\label{sec-back}

Social determinants of health (SDoH) are non-clinical factors influencing health, functioning, and quality of life outcomes and risks. For example, SDoH factors include where people are born, live, learn, work, play, worship, and their age~\cite{world2008social, marmot2012european, gucciardi2014intersection}. Decades of studies have shown that medical care accounts for only 10–20\% of an individual's health status. However, social, behavioral, and genetic factors also significantly influence health risks, outcomes, access to health services, and adherence to prescribed care~\cite{singh2010neighborhood,yang2022examining}. Thus, addressing SDoH is critical for increasing healthcare quality, decreasing health disparities, and informing clinical decision-making~\cite{koh2011healthy}.

Unfortunately, Electronic Health Records (EHRs) do not generally code SDoH information in structured data, e.g., not in ICD-10 codes~\cite{quan2005coding}. Instead, healthcare organizations and professionals typically record SDoH in unstructured narrative clinical notes. Thus, this critical patient information is not easily accessible. Healthcare practitioners need to translate them into structured data to support downstream secondary use applications, like disease surveillance and clinical decision support~\cite{karran2020low}. Traditionally, medical practitioners have to manually collect information from unstructured data, such as medical records, in order to make diagnoses and treatment plans. This process, known as medical record review, can be challenging and time-consuming. The extensive paperwork burden can increase fatigue, reduce job satisfaction, and contribute to medical errors and adverse events~\cite{conway2019moonstone}. Automating the extraction of SDoH from unstructured clinical notes using Natural Language Processing (NLP) techniques can help to reduce the worklofad for medical practitioners, improve the accuracy and efficiency of the information collection process, and generate a comprehensive representation of the patient about their social, behavioral, and environmental information for downstream tasks. This approach has been shown to be effective in previous research, as demonstrated in prior work~\cite{jensen2012mining, hatef2019assessing, lybarger2021annotating}.

Previous studies on leveraging NLP to automate the extraction of SDoH information has included lexicons/rule-based methods~\cite{hatef2019assessing,bejan2018mining} and deep learning approaches~\cite{feller2020detecting,stemerman2021identification,lybarger2021annotating}.
In this work, we introduce a novel system with three main components to extract event-based SDoH information from clinical notes: named entity recognition (NER)~\cite{sang2003introduction}, relation extraction (RE)~\cite{zelenko2003kernel} and text classification (TC)~\cite{garla2012ontology}. We use the Social History Annotation Corpus (SHAC) developed for the 2022 N2C2 Shared Task---which is based on the work by~\Mycite{lybarger2021annotating}. One of the main challenges in extracting SDoH from text is a large number of overlapping entities. For example, \Mycite{lybarger2021annotating} defines smoking status as a SDoH. In their corpus, the span of text ``2-3 cig per day'' includes four entities: the StatusTime argument (``2-3 cig per day''), the Amount argument (``2-3 cig''), the frequency (``per day``), and the type (``cig''). Even worse, entities with the exact same spans can refer to two separate entities. For example, ``marijuana'' represents the entity Drug, and it represents the entity Type (i.e., because it refers to a type of drug) in the \Mycite{lybarger2021annotating} corpus.

Recently, several methods have been proposed for handling overlap in named entity recognition (NER) tasks~\cite{sohrab-miwa-2018-deep, wang2019combining, zhong2021frustratingly, yuan2022fusing}. Some papers have designed different tagging schemes~\cite{wang2020pyramid, strakova2019neural} by combining token-level classes to deal with overlapping NER, which may cause data sparsity issues (e.g., a word can be labeled as B-ORG-I-PER if it is the start of an organization span and the inner part of a person's name). However, if two entity types overlap infrequently, this can cause a data-sparsity issue. Span-based models are another approach for handling overlapping entities~\cite {yan2021unified, huang2022extract}. These models follow a two-stage framework, first extracting all possible text spans from the text and then using filters to reduce the total search space and computational complexity~\cite {yan2021unified, huang2022extract}. \Mycite{rojas2022simple} shows that simply training an individual model for every entity type (assuming overlapping entities only appear across entity types) produces better performance than more complex prior methods. However, training a single model for every entity type can be wasteful regarding memory usage. Moreover, if the number of entity types is large, the deployment of many models can be difficult.

To address limitations in prior work for extracting SDoH information from text using the NER approach, we propose a unified marker-based sequence labeling model for the simultaneous extraction of triggers and arguments in a single NER model. This model is then used as part of a larger event extraction system, which outperforms recent methods introduced in the 2022 N2C2 shared task. Our method is inspired by the success of prefix-based prompt-learning~\cite{soares2019matching, zhong2021frustratingly, huang2022extract} and the work by \Mycite{rojas2022simple} that shows individual models for each entity outperforming more complex overlapping NER systems. Intuitively, our approach simulates individual models trained for every entity type into a single system. 


\begin{figure}[t]
    \centering
    \includegraphics[width=0.99\linewidth]{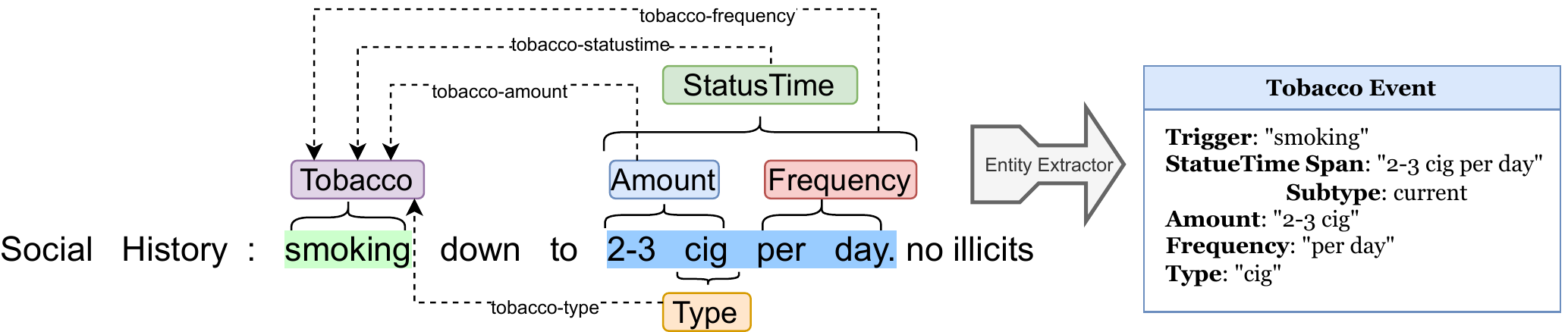}
    \caption{An example of Social Determinants of Health extraction.}
    \label{fig:sdoh-structure}
\end{figure}

In summary, this paper makes the following contributions:
\begin{enumerate}

\item We propose a simple yet novel system for SDoH information extraction. Our system achieves state-of-the-art performance compared to other competitive systems submitted to the National NLP Clinical Challenges (n2c2) shared task.

\item We propose a novel marker-based sequence labeling method for extracting all possible triggers and argument entities while handling overlap. The method is shown to outperform more complex methods developed for overlapping NER.

\item We conduct an ablation-like analysis to understand which components of our system have the greatest potential for improving SDoH extraction. Moreover, we perform an error analysis to provide future avenues of research.
\end{enumerate}

\section{METHODOLOGY}

The goal of the SDoH extraction task is to extract ``triggers'' and ``arguments''. Triggers are mentions of SDoH factors (e.g., Alcohol, Drug, Tabacco, Living Status, and Employment). Arguments link to the triggers to provide further context. An example is provided in Figure~\ref{fig:sdoh-structure}. The trigger extracted is ``smoking'' which was assigned the trigger entity Tobacco. Next, there are four argument entities extracted: StatusTime, Amount, Frequency, and Type. Note that these entities can be nested (overlapping), as discussed in the Background and Significance Section. Intuitively, the argument entities provide information about the trigger entity, e.g., what they were smoking and how often they smoked. Some arguments (e.g., StatusTime) are also classified into specific subtypes to provide a standardized format for important information. In this case, we see that the person is a ``current'' smoker. Overall, while our main methodological advances come from the NER component (which we justify via a careful analysis), each piece works together to extract SDoH information. Finally, we describe the exact entity types for triggers, arguments, and all of the subtypes in their respective subsections below.

\begin{figure}[t]
    \centering
    \includegraphics[width=0.9\linewidth]{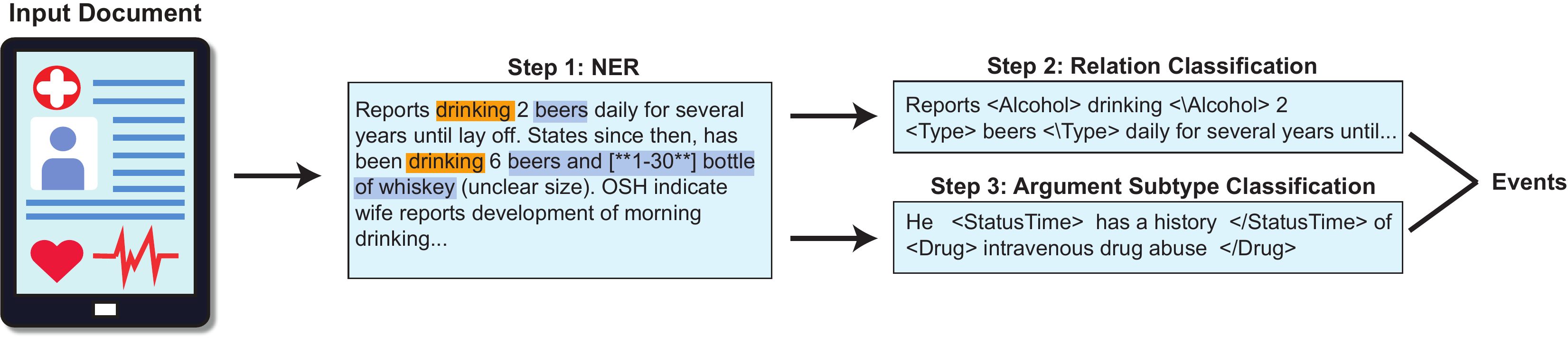}
    \caption{Overview of Marker-Based Pipeline.}
    \label{fig:pipeline}
\end{figure}

\subsection{Named Entity Recognition.} 

The first stage of our SDoH system is to extract all trigger and argument entities within the text. There are five unique trigger entities: Drug, Alcohol, Tobacco, Employment, and LivingStatus. Likewise, there are nine unique argument entity types: StatusTime, StatusEmploy, TypeLiving, Type, Method, History, Duration, Frequency, and Amount. Every argument type does not match every trigger type. For instance, TypeLiving refers to text spans that mention how a person lives (e.g., whether they are homeless), which is not directly applicable to the other triggers such as Drug and Employment.

Formally, we frame this as a traditional NER sequential labeling task, where a sequence $S$ consisting of n tokens $w_1, w_2, \ldots, w_n$, where $n$ denotes the length of the sequence is classified into a sequence of labels $L$ defined as $l_1, l_2, \ldots, l_n$. Specifically, we model
\begin{equation*}
    P(l_1, \ldots, l_n | w_1, \ldots, w_n)
\end{equation*}
where each label $l_i$ represents an entity type in Beginning-Inside-Outside (BIO) format (e.g., B-Drug, I-Drug,  B-Type)~\cite{ramshaw1999text}. Outside, or O, represents a token not classified into one of the SDOH trigger or argument entities. This traditional approach does not handle overlapping entities. In the SDOH corpus, overlapping entities appear across entity types. Generally, an entity does not overlap with an entity of the same type. This assumption is also used in prior overlapping entity work~\cite{rojas2022simple}. However, to overcome this prior work, \Mycite{rojas2022simple} trains an independent classifier for every entity type. For instance, a single model would predict all Drug entities, while another model would be dedicated to Employment. This approach could result in 14 unique models in our corpus (i.e., a model for each trigger and argument entity type), e.g., $P_{Drug}(l_1, \ldots, l_n | w_1, \ldots, w_n)$, $P_{alcohol}(l_1, \ldots, l_n | w_1, \ldots, w_n)$, etc. Moreover, there may be information about one entity that can help improve the prediction of another. However, using independent models will overcome the issue of overlapping entities, but it will also cause the loss of access to cross-entity information.

\begin{figure}[t]
    \centering
    \includegraphics[width=0.9\linewidth]{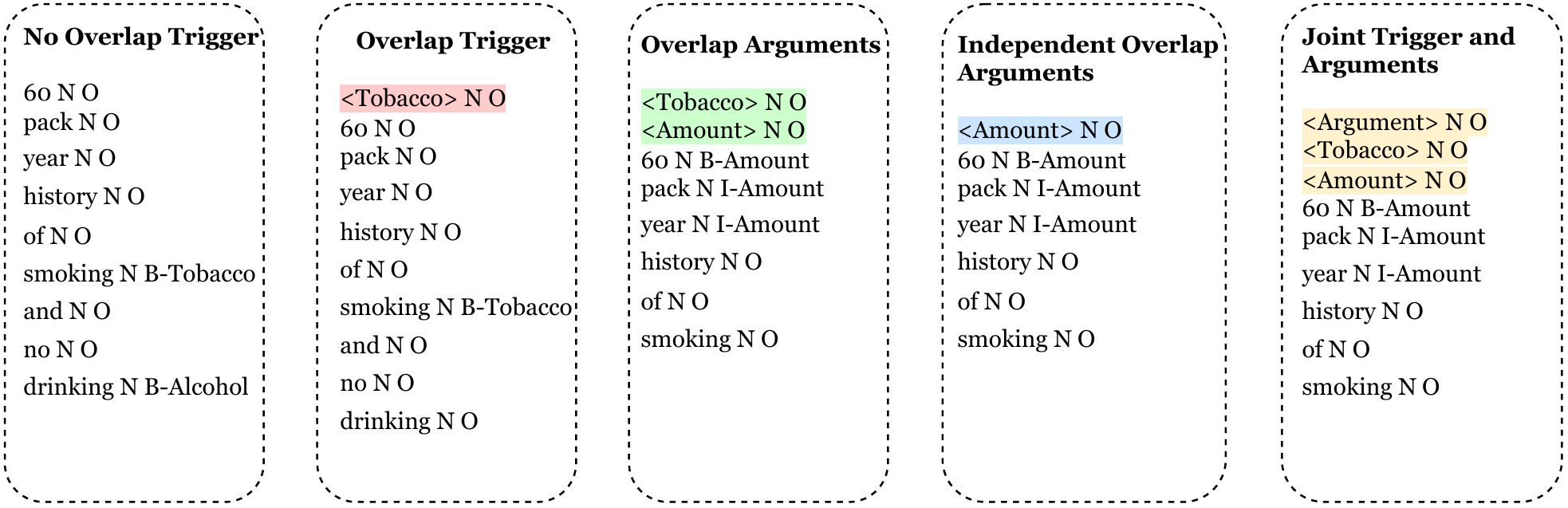}
    \caption{Examples of the markers used for each of our NER systems.}
    \label{fig:marker-schema}
\end{figure}

To overcome the limitations of training a separate model for each entity type, we explore methods of handling overlap without training multiple models by exploring different types of entity type markers, which have been shown to be effective at injecting information into the model~\cite{huang2022extract,van2021crosslingual, hsu2022degree}. We explore two unique methods of training a joint NER model for the trigger entities, two models for the arguments, and one joint model for triggers and arguments. The summary of each marker-based system for each variation is shown in Figure~\ref{fig:marker-schema}. We describe each model below. 

\paragraph{Trigger Model 1 (No Overlap Trigger).} First, for triggers, we explore the use of the traditional flat NER, where we ignore overlap between trigger entities. We found that there is not a substantial amount of overlap between trigger entities, though it does appear within the dataset. Specifically, given the input sentence, we will simultaneously predict all trigger entity types by modeling.
\begin{equation*}
        P^t(l_1, \ldots, l_n | w_1, \ldots, w_n)
\end{equation*}
where $P^t()$ represents the NER model for all triggers. Each token will be assigned one, and only one, BIO formatted label $l_i$.

\paragraph{Trigger Model 2 (Overlap Trigger).} Next, we explore a trigger model that can handle overlap. Intuitively, we simulate training a single model for every trigger entity type using a marker $k$. Intuitively, instead of predicting all trigger entity types in a single pass of the sentence and, thus, only assigning a single class to each token, we make predictions by first conditioning on the entity type we want to predict. Formally, we model
\begin{equation*}
        P^t(l_1, \ldots, l_n | w_1, \ldots, w_n, k)
\end{equation*}
which will only make predictions for each token $w_1, \ldots, w_n$ for trigger $k$ or not $k$. As we change $k$, the predictions will change. We implement this model by prepending a trigger type marker $k$ to the start of each sequence that is formatted as $<$TriggerName$>$ (e.g., $<$Tobacco$>$), which transforms a sequence of tokens $w_1, \ldots, w_n$ to $k, w_1, \ldots, w_n$. An example is provided in Figure~\ref{fig:marker-schema}.

\paragraph{Argument Model 1 (Independent Overlap Arguments).} For the arguments, there is substantial overlap between entities. Hence, completely ignoring overlap is not feasible. The first argument model we explore involves training an Overlapping Argument model for each trigger. Specifically, we train a model similar to Overlap Trigger for arguments, but the model is trained for each trigger's arguments. For instance, train a model for all of the Tobacco trigger's arguments, StatusTime, Amount, Frequency, and Type. Likewise, we do the same for the other triggers, resulting in five models. Formally, we train a model
\begin{equation*}
        P^a_k(l_1, \ldots, l_n | w_1, \ldots, w_n, q)
\end{equation*}
where $q$ represents an argument for trigger $k$. Similar to the Overlap Trigger model, we implement this by prepending the marker $q$ to the sequence of tokens $w_1, \ldots, w_n$ to form $q, w_1, \ldots, w_n$. Again, at inference time, we only predict one entity type $q \in Q$ where $Q$ is the set of arguments for trigger $k$. To generate a different argument entity, we change $q$ (e.g., we prepend $<$Type$>$ to predict the Type argument and $<$Frequency$>$ to predict the frequency entity).

\paragraph{Argument Model 2: Overlap Arguments.} Instead of learning a joint argument model across all five triggers, we also experiment with a single argument model across all triggers. Formally, we model
\begin{equation*}
        P^a(l_1, \ldots, l_n | w_1, \ldots, w_n, k, q)
\end{equation*}
which conditions on trigger $k$ and argument $q$. Again, we implement this by prepending both an argument and trigger marker, transforming the tokens $w_1, \ldots, w_n$ to $k, q, w_1, \ldots, w_n$.

\paragraph{Joint Triggers and Arguments.} The final model we explore is a single joint model for Triggers and Arguments. Note that there is a substantial overlap between trigger and argument entities. Hence, this joint model tests the complete ability to handle the overlap of our marker-based system. This model is an extension of the Overlap Trigger and Overlap Arguments models. Specifically, we change what is prepended depending on what should be predicted. Formally, we model
\begin{equation*}
        P(l_1, \ldots, l_n | w_1, \ldots, w_n, k, q, z)
\end{equation*}
where $k$ is a trigger marker (e.g., $<$Drug$>$), $q$ is an argument marker (e.g., $<$Type$>$), and $z$ is a marker that indicates whether we should predict a trigger or argument (e.g., $<$Trigger$>$ or $<$Argument$>$). If we are predicting a trigger, then $q$ is set to the empty string. Specifically, if we predict a trigger, $w_1, \ldots, w_n$ to $m, k, q, w_1, \ldots, w_n$. As an example, if we want to predict trigger Drug entities, we would modify the input sequence to start with ``$<$Trigger$>$ $<$Drug$>$''. To predict different entities, we modify the inputs to the system as appropriate.

\paragraph{Combinations.} In our experiments, we explore five combinations of the models above: ``No Overlap Trigger + Ind. Overlap Arguments'', ``No Overlap Trigger + Overlap Arguments'', ``Overlap Trigger + No Overlap Arguments'', ``Overlap Trigger + Overlap Arguments'', and ``Joint Trigger and Arguments''.


\subsection{Relation Classification (RC).} 

\begin{figure}[t]
    \centering
    \includegraphics[width=0.9\linewidth]{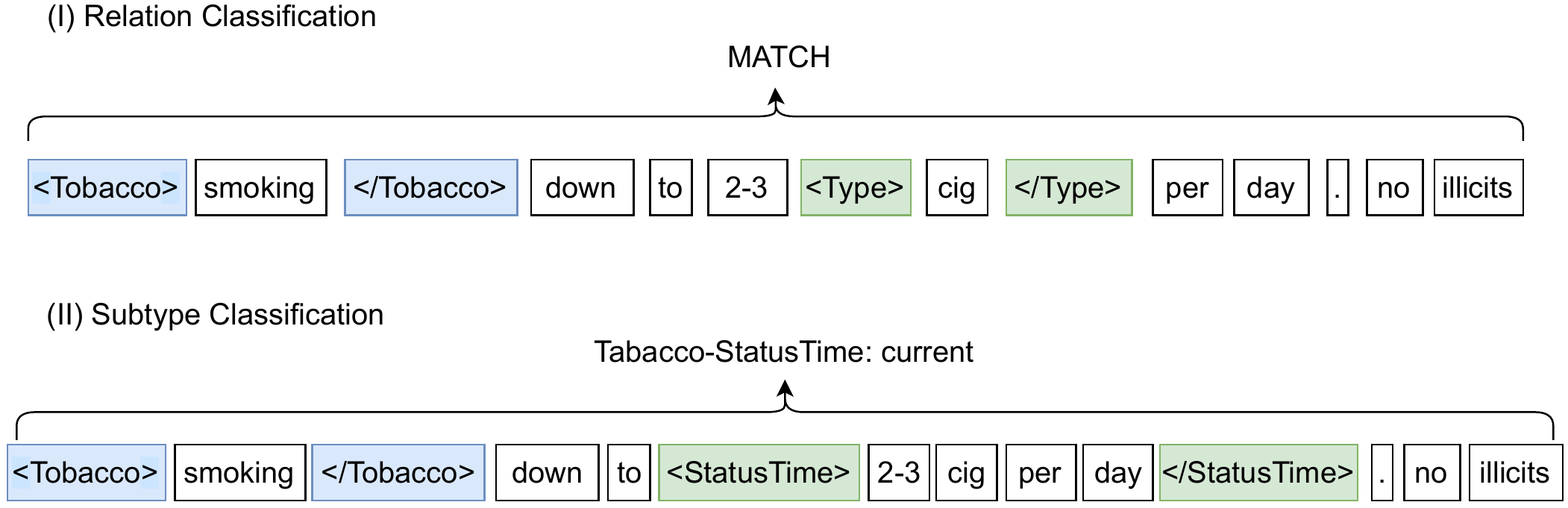}
    \caption{Examples for the relation classification and subtype classification. (I) Relation Classification is a binary classification task that determines whether a relation exists between trigger and argument. The two possible classes are ``match'' and ``not match''. (II) For subtype classification, a labeled argument is classified into one of several pre-defined subtypes, where each has a specific semantic meaning (e.g., ``current'' drug user). }
    \label{fig:classification}
\end{figure}

In our models for NER, we can map an extracted argument to a trigger of the correct type. However, there may be multiple triggers of the same type (e.g., multiple Alcohol types in Figure~\ref{fig:pipeline}). Therefore, matching arguments to an associated trigger instance is not possible with the NER models alone. Hence, we propose a relation classification framework to match arguments to their respective triggers. To follow a similar framework as our marker-based NER system, we applied the traditional RC (Matching the Blanks) approach~\cite{baldini-soares-etal-2019-matching,lee2022bertsrc, zhang2019ernie,peters2019knowledge, zhong2021frustratingly}. Specifically, we model the probability that an argument should map to a trigger as
\begin{equation*}
    P(y = \text{match}\, |\, w_1, \ldots, w_n , e_1, e_2)
\end{equation*}
where $e_1$ represents the trigger entity and $e_2$ represents the argument entity. We model this classification task by wrapping the entities with markers. For example, given the sentence ``smoking down to 2-3 cig per day'', if we want to check if the type argument ``cig'' maps to the trigger ``smoking'', then the text is modified as ``$<$Tobacco$>$ smoking $</$Tobacco$>$ down to 2-3 $<$Type$>$ cig $</$Type$>$ per day''. See another example in Figure~\ref{fig:classification}.

\subsection{Argument Subtype Classification.} 


The final piece of our SDOH extraction framework involves subtype classification. There are arguments (e.g., Employment Status) that provide important information. However, it is generally stated in a wide array of formats. For instance, ``John was just laid off work'' and ``John is not working'' are both mention that a person is unemployed. There are six arguments that are categorized into subtypes: Alcohol StatusTime, Drug StatusTime, Tobacco StatusTime, Employment Status, LivingStatus StatusTime, and LivingStatus TypeLiving. Each StatusTime subtype can take one of three categories: current, past, and future. Employment Status can be employed, unemployed, retired, on disability, student, or homemaker. LivingSTatus TypeLiving can be alone, current, and past. 

To detect subtypes, we use a similar framework as our relation classification component. Specifically, we model
\begin{equation*}
    P(s |\, w_1, \ldots, w_n , e_1, e_2)
\end{equation*}
where $e_2$ represents the status argument we are subtyping and $e_1$ is its respective trigger entity matched via the relation classification model. $s$ represents the subtype. Again, we model this via markers within the text, just like the relation classification task. For instance, given the sentence, ``smoking down to 2-3 cig per day'', the StatusTime argument and Tobacco trigger are marked as ``$<$Tobacco$>$ smoking $</$Tobacco$>$ down to $<$StatusTime$>$ 2-3 cig per day $</$StatusTime$>$'', where the correct subtype would be ``current''. We train a single model to capture all subtypes across the six arguments. See another example in Figure~\ref{fig:classification}.

\subsection{Implementation Details} 

For the NER models, we train a Bi-directional Long Short-Term Memory network (BiLSTM) with conditional random fields~\cite{luo2018attention}. For the model, we explore two types of input embeddings: Flair~\cite{akbik2019flair} and T5-3B~\cite{2020t5}. For the Flair embedding model, we trained a marker-based NER model using a sample dropout of 0.4, a hidden layer size of 128, a learning rate of 0.1, and 25 epochs with a mini-batch size of 16. We save the model after each epoch and use the best version based on the validation dataset. The T5-3B embedding model was trained in a similar fashion, with the exception of a sample dropout of 0.3, a hidden layer size of 1024, a maximum of 15 epochs, and a learning rate of 0.025. Both models fine-tuned the embedding layers. All NER models were implemented using the Flair software framework developed by~\Mycite{akbik2019flair}.~\footnote{\url{https://github.com/flairNLP/flair}} For the relation classification and subtype classification models, we use a RoBERTa-base model~\cite{liu2019roberta} with an Adam optimizer~\cite{kingma2014adam} and the CosineAnnealingLR scheduler~\cite{loshchilov2016sgdr}, a learning rate of 1e-5, and train for a maximum of 20 epochs. Again, the best epoch is chosen using the validation data. Finally, all experiments were performed on four NVidia GeForce GTX 1080 Ti GPUs and one NVidia A6000.

\begin{table}[t]
\centering
\begin{tabular}{llrrr}
\toprule
\multicolumn{1}{c}{\textbf{Dataset}} & \textbf{Subset} & \textbf{\# of Documents}  & \textbf{Max Words} & \textbf{AVG Words} \\ \midrule
\multirow{3}{*}{\textbf{MIMIC-III}} & \textbf{Train} & 1316 & 229 & 65.34 \\
 & \textbf{Dev} & 188 & 82 & 44.34 \\
 & \textbf{Test} & 373 & 192 & 44.50  \\ \midrule
\multirow{3}{*}{\textbf{UW}} & \textbf{Train} & 1751 & 437 & 54.22 \\
 & \textbf{Dev} & 259 & 99 & 37.47 \\
 & \textbf{Test} & 518 & 288 & 37.16 \\ \bottomrule
\end{tabular}
\caption{Dataset statistics for the MIMIC-III and UW datasets. Statistics include the number of examples/documents in each subset, max words in a document, and the average words per document.\vspace{0em}}
\label{tab:stats}
\end{table}

\section{EXPERIMENTAL RESULTS}

In this section, we describe the data, evaluation metrics, and report results, and an error analysis. 

\subsection{Datasets}  

We conducted our experiments on the 2022 N2C2 shared task version of the SHAC~\cite{lybarger2021annotating} corpora. The dataset consists of 4480 annotated social history sections (70\% train, 10\% development, 20\% test) from MIMIC-III and the University of Washington Harborview Medical Centers data (UW). The systems are evaluated for three scenarios. First, \textbf{Task A} involves training and evaluating on the MIMIC-III data (i.e., MIMIC-III $\rightarrow$ MIMIC-III). \textbf{Task B} measures generalizability which involves training on the MIMIC-III and evaluating on UW data (i.e., MIMIC-III $\rightarrow$ UW). Finally, \textbf{Task C} involves training on MIMIC-III and UW data and evaluating on UW data (i.e., MIMIC-III + UW $\rightarrow$ UW). Table~\ref{tab:stats} presents basic information about the datasets.

\subsection{Evaluation Metrics}

Performance is evaluated using the following metrics: overall precision (P), recall (R), and F1-score (F1), which is a micro-average of all trigger types, argument types, and argument subtypes (i.e., true positives, false positives, and false negatives are summed across all categories). In all of our analysis, we use the evaluation tools provided by the N2C2 shared task organizers~\footnote{\url{https://github.com/Lybarger/brat_scoring}}.

\subsection{Overall Results}
Table~\ref{tab:competition} shows the overall performance of our systems compared to the best models in the 2022 N2C2 shared task. Although our model is simple, the marker-based system approach outperforms prior work. Specifically, our system with flair word embeddings and joint NER model achieves similar performance to the best performing systems without using an ensemble or manually curated rules (.8829 vs .9008 for Task A and .8788 vs .8886 for Task C). Moreover, our results are further improved using a larger pre-trained model T5-3B. The T5-3B embeddings and Joint Trigger + Argument NER model achieves an absolute F1 score improvement compared with best competition results of .0104, .0354, and .0156 for Task A, Task B, and Task C, respectively. The improvements demonstrate the effectiveness of prefixing entity type marker in front of sentence to handle overlapping NER. We also find that using joint models generally outperforms using more models. For example, Flair + RoBERTa Joint Trigger and Argument has an F1 of .7523 for Task B, while Flair + RoBERTa No Overlap Trigger + Ind. Overlap Argument has an F1 of of .7189. One possible reason for the excellent performance is that when we train joint models, more cross-entity information is shared, similar to what happens with multi-task learning.

\begin{table}[t]
\resizebox{\linewidth}{!}{
\begin{tabular}{@{}llrrrrrrrrr@{}}
\toprule
\multicolumn{1}{l}{} & \multicolumn{1}{l}{} & \multicolumn{3}{c}{\textbf{Task A}} & \multicolumn{3}{c}{\textbf{Task B}} & \multicolumn{3}{c}{\textbf{Task C}} \\ \cmidrule(lr){3-5} \cmidrule(lr){6-8} \cmidrule(lr){9-11}
\textbf{Representations} & \textbf{NER Method} & \textbf{P} & \textbf{R} & \textbf{F1} & \textbf{P} & \textbf{R} & \textbf{F1} & \textbf{P} & \textbf{R} & \textbf{F1} \\ \midrule
\multicolumn{1}{l}{} & \textbf{Best Competition Models} & .9093 & .8925 & .9008 & .8108 & .7400 & .7738 & .8906 & .8867 & .8886 \\ \midrule
\cellcolor[HTML]{FFFFFF} & \textbf{Joint Trigger and Argument} & \textbf{.9073} & .8597 & \textbf{.8828} & \textbf{.8016} & \textbf{.7088} & \textbf{.7523} & \textbf{.8926} & .8642 & \textbf{.8781} \\
\cellcolor[HTML]{FFFFFF} & \textbf{Overlap Trigger + Overlap Argument} & .9010 & \textbf{.8655} & \textbf{.8829} & .7967 & .7036 & .7473 & .8837 & \textbf{.8741} & \textbf{.8788} \\
\cellcolor[HTML]{FFFFFF} & \textbf{Overlap Trigger + Ind. Overlap Argument} & .9001 & .8643 & .8818 & .7856 & .7040 & .7425 & .8870 & .8650 & .8759 \\
\cellcolor[HTML]{FFFFFF} & \textbf{No Overlap Trigger + Overlap Argument} & .8915 & .8594 & .8752 & .7835 & .6714 & .7231 & .8707 & .8677 & .8692 \\
\multirow{-5}{*}{\cellcolor[HTML]{FFFFFF}\textbf{Flair + RoBERTa}} & \textbf{No Overlap Trigger + Ind. Overlap Argument} & .8890 & .8580 & .8732 & .7733 & .6717 & .7189 & .8739 & .8581 & .8659 \\ \midrule
\cellcolor[HTML]{FFFFFF} & \textbf{Joint Trigger and Argument} & .9035 & \textbf{.9167 }& \textbf{.9101} & .8144 & .7964 &  \textbf{.8053} & .9002 & \textbf{.9049} & \textbf{.9025} \\
\cellcolor[HTML]{FFFFFF} & \textbf{Overlap Trigger + Overlap Argument} & \textbf{.9132} & .9092 & \textbf{.9112} & \textbf{.8194} & \textbf{.7992} &  \textbf{.8092} & \textbf{.9036} & \textbf{.9049} & \textbf{.9042} \\
\cellcolor[HTML]{FFFFFF} & \textbf{Overlap Trigger + Ind. Overlap Argument} & .9036 & .9020 & .9028 & .8029 & .7800 & .7913 & .8982 & .9005 & .8994 \\
\cellcolor[HTML]{FFFFFF} & \textbf{No Overlap Trigger + Overlap Argument} & .9009 & .8980 & .8994 & .8165 & .7780 & .7968 & .8969 & .9049 & .9009 \\
\multirow{-5}{*}{\textbf{T5-3B + RoBERTa}} & \textbf{No Overlap Trigger + Ind. Overlap Argument} & .8924 & .8914 & .8919 & .8014 & .7575 & .7788 & .8916 & .9009 & .8962 \\ \bottomrule
\end{tabular}}

\caption{Overall Performance across the three tasks: Task A (MIMIC $\rightarrow$ MIMIC), Task B (MIMIC $\rightarrow$ UW), Task C (MIMIC+UW $\rightarrow$ UW). Best scores are \textbf{bolded} for the best model(s) for each set of embedding types (Flair + Roberta and T5-3B + RoBERTa).}
\label{tab:competition}
\end{table}

\subsection{Analysis of System Component Importance}
There are three major components to our SDoH extraction system: NER, relation classification, and subtype classification. For future work, which piece can provide the most benefit if improved?
To understand each component better, we run an ablation-like experiment where we replace each component with the ground-truth predictions. Intuitively, we are trying to understand if we improved a single component, which has the most \textit{potential} impact on the entire system.  Table~\ref{tab:ablation-taskac} shows the results of study for Task A and Task C. By comparing, we find that using ground truth for argument-level NER yields the largest potential improvement (.0433 for Task A and .0403 fot Task C). The next largest potential improvement comes from the Relation Classfication model. The component with the lowest potential impact on the overall performance is subtype classification, with an improvement of .0193 for Task A and .0162 for Task C. 

\begin{table}[t]
\centering
\resizebox{.8\linewidth}{!}{
\begin{tabular}{@{}lrrrrrrrr@{}}
\toprule
 & \multicolumn{4}{c}{\textbf{Task A}} & \multicolumn{4}{c}{\textbf{Task C}} \\ \cmidrule(lr){2-5} \cmidrule(lr){6-9}
\multicolumn{1}{c}{\textbf{Model}} & \multicolumn{1}{c}{\textbf{P}} & \multicolumn{1}{c}{\textbf{R}} & \multicolumn{1}{c}{\textbf{F1}} & \multicolumn{1}{c}{\textbf{Diff F1}} & \multicolumn{1}{c}{\textbf{P}} & \multicolumn{1}{c}{\textbf{R}} & \multicolumn{1}{c}{\textbf{F1}} & \multicolumn{1}{c}{\textbf{Diff F1}}\\ \midrule
\textbf{Joint Trigger and Argument} & .8994 & .9074 & .9034 & --- & .9064 & .9189 & .9126 & --- \\
\textbf{+ Perfect NER-Trigger} & .9410 & .9207 & .9308 & .0274 & .9391 & .9363 & .9377 & .0251 \\
\textbf{+ Perfect NER-Argument} & .9482 & .9451 & .9467 &  \textbf{.0433} & .9579 & .9480 & .9529 &  \textbf{.0403} \\
\textbf{+ Perfect Subtype Classification} & .9186 & .9268 & .9227 & .0193 & .9225 & .9352 & .9288 & .0162 \\
\textbf{+ Perfect Relation Classification} & .9639 & .9046 & .9333 & .0300 & .9671 & .9152 & .9404 & .0278 \\ 
\bottomrule
\end{tabular}}

\caption{Analysis of system component importance for Task A and C using their respective development sets. The biggest differences are \textbf{bolded}. \vspace{0em}}
\label{tab:ablation-taskac}
\end{table}

\subsection{Comparison to a State-of-the-art Span-based model}
As mentioned in the Background and Significance Section, there has been significant progress in developing complex overlapping. While some research has shown that training independent models outperform many of the recent methods~\cite{rojas2022simple}, it is important to compare them as a baseline. Hence, we applied a recent span-based method Triaffine~\cite{yuan2022fusing} to using publicly available source code on the N2C2 shared task data~\footnote{\url{https://github.com/GanjinZero/Triaffine-nested-ner}}. This approach allows the model to capture complex dependencies and interactions between different elements in the input text, potentially improving its performance on tasks such as overlapped named entity recognition. Triaffine is currently a state-of-the-art method in this area~\cite{yuan2022fusing}. We compare two versions of the model, one that trains triggers and arguments jointly (Joint Trigger + Argument) and one that trains a model for the triggers separately from the arguments (Independent Trigger + Argument). We report the results in Table~\ref{tab:triaffine}. Overall, we find that the independent model substantially outperforms the joint model across all three tasks (e.g., .8594 vs. .5942 for Task A). The reason the Joint model suffers is that it is not capable of handling cases where the triggers overlap exactly with the span of an argument. Our method is capable of handling this by predicting each entity one at a time using markers. We also compare with assuming a perfect relation classification because the span-based model does not have information about matches between arguments and trigger types. Our models contain this information by including a marker for the trigger and the argument for argument prediction. Yet, even with a perfect relation classification model, it still underperforms our best approach without a perfect model.

\begin{table}[t]
\centering
\resizebox{0.99\linewidth}{!}{
\begin{tabular}{@{}lrrrrrrrrr@{}}
\toprule
 & \multicolumn{3}{c}{\textbf{Task A}} & \multicolumn{3}{c}{\textbf{Task B}} & \multicolumn{3}{c}{\textbf{Task C}}  \\ \cmidrule(lr){2-4} \cmidrule(lr){5-7} \cmidrule(lr){8-10}
\multicolumn{1}{c}{\textbf{Model}} & \multicolumn{1}{c}{\textbf{P}} & \multicolumn{1}{c}{\textbf{R}} & \multicolumn{1}{c}{\textbf{F1}} & \multicolumn{1}{c}{\textbf{P}} & \multicolumn{1}{c}{\textbf{R}} & \multicolumn{1}{c}{\textbf{F1}} & \multicolumn{1}{c}{\textbf{P}} & \multicolumn{1}{c}{\textbf{R}} & \multicolumn{1}{c}{\textbf{F1}} \\ \midrule
\textbf{Triaffine: Independent Trigger + Argument} & \textbf{.9050 }& .8182 & .8594 & .7889 & .6641 & .7211 & .8876 & .8462 & .8664  \\
\multicolumn{1}{r}{\textbf{+ Perfect Relation Classification}} & {.9561}	& .8222	  & .8841 & .8108	& .7400	 & .7738 & .8906	& .8867	 & .8886  \\ \cmidrule(lr){2-10}
\textbf{Triaffine: Joint Trigger + Argument} & .8585 & .4543 & .5942 & \textbf{.8326}	& .4377 & .5738 & \textbf{.9101} & .5555 & .6899 \\ 
 \midrule
\textbf{T5-3B + RoBERTa Joint Trigger + Argument (ours)} & .9035 & .\textbf{9167} &  \textbf{.9101}& .8144 & .7964 & \textbf{.8053} & .9002 & \textbf{.9049} & \textbf{.9025} \\ \bottomrule
\end{tabular}}
\caption{Comparison to the Triaffine~\cite{yuan2022fusing} span-based model for overlapping entities. Results are on the test data. The largest numbers are \textbf{bolded}. \vspace{0em}}
\label{tab:triaffine}
\end{table}

\subsection{Error Analysis}

We analyze common errors made by our Joint Trigger and Argument model. First, when there are direct mentions of different (unique) types of drugs that have different StatusTime (e.g., current vs. past), annotators will label each as separate triggers. For instance,
\begin{center}
\vspace{.75em}
\noindent\fbox{%
    \parbox{.93\linewidth}{%
        ``\textbf{Illicit drugs}:  current \textit{marijuana use}, \textit{cocaine} quit 5 years ago.''
    }
}
\vspace{.75em}
\end{center}
 has two Drug triggers: ``marijuana use'' and ``cocaine''. Yet, our model only predicts the more general ``Illicit drugs'' as the trigger entity. We hypothesize that our model does not differentiate general concepts (e.g., ``Illicit drugs'') from more specific instances of the concept (e.g., ``marijuana'' and ``cocaine''. This is because it is not modeled explicitly in the architecture, moreover, the data general contains more instances of the generic mentions than the more specific mentions. Another example of this is found in the example
\begin{center}
\vspace{.75em}
\noindent\fbox{%
    \parbox{.93\linewidth}{%
        ``She \textbf{drinks} 2-3 \textit{alcoholic} beverages per week.''
    }
}
\vspace{.75em}
\end{center}
where our model predict ``drinks'' as trigger, while the ground truth is ``alcoholic''. Based on the criteria~\Mycite{lybarger2021annotating}, the phrase describing general substance (i.e. alcohol, tobacco, or drug) or substance-related verb, such as drink can be a trigger. When both appear, the more specific concept should be used. Yet, again, our model fails to understand this underlying semantic meaning and does not differentiate instances from generic types. This error is  very common for other trigger types. For example, our model incorrectly predict ``smokes'' as trigger instead of the ground-truth ``cigarettes'' often.  Likewise, for the employment trigger, our model will predict ``worked'' as trigger instead of ``retired'' in some examples. Another common error type happens for uncommon noun phrases. For instance, in the example
\begin{center}
\vspace{.75em}
\noindent\fbox{%
    \parbox{.93\linewidth}{%
       ``Currently at \textit{a rehab facility}, but previously living with his wife at home.''
    }
}
\vspace{.75em}
\end{center}
the ground-truth for the LivingStatus trigger is ``a rehab facility'', but our model fails to detect it. Another example of this error type includes
\begin{center}
\vspace{.75em}
\noindent\fbox{%
    \parbox{.93\linewidth}{%
        ``Works in \textbf{finance} \textit{at Mass Eye \& Ear}.''
    }
}
\vspace{.75em}
\end{center}
where our model predicts the Type argument for the Employment trigger as ``finance'', while the ground-truth is ``finance at Mass Eye \& Ear''. Again, this indicates our models strugles with novel noun phrases, particularly when they include prepositional phrases.
A future interesting research avenue would explore methods for incorporating external knowledge bases into transformer models. This could potentially help the model make more accurate predictions and avoid errors. One way to incorporate external knowledge into transformer models is through the use of external memory networks, which have been shown to be effective at incorporating common sense into language models~\cite{xing2021km}. 

\section{CONCLUSION}
In this paper, we present our approach for extracting SDoH events from clinical notes using the N2C2-2022 Task 2 shared task dataset. We introduce a novel NER system to extract overlapped entities and propose a multiple pipeline system to extract SDoH events, including NER, Relation Classification, and Subtype Classification models, that results in a new state-of-the-art performance for the N2C2 data. In future efforts, we aim to enhance our NER model by utilizing structured knowledge bases through demonstration-based learning~\cite{lee2022good}, such as providing the sentence of task demonstrations or entity type descriptions instead of just using simple entity type markers for in-context learning. This can easily be integrated into our framework, and we hypothesize it would help low-resource entities.



\section*{FUNDING}
 
 This material is based upon work supported by the National Science Foundation (NSF) under Grant~No. 2145357.

\section*{COMPETING INTERESTS}

None

\section*{CONTRIBUTORS}

Xingmeng Zhao performed the experiments and drafted the initial manuscript. Anthony Rios conceived of the study, oversaw the design, and reviewed and approved the manuscript.


\newpage 
\bibliographystyle{unsrtnat}
\bibliography{scoonerdb.bib}
\end{document}